\pdfoutput=1

\documentclass[11pt]{article}

\usepackage[]{acl}

\usepackage{times}
\usepackage{latexsym}

\usepackage{graphicx} 
\usepackage{multirow} 
\usepackage{booktabs} 
\usepackage{amssymb}

\newcommand{\legentlogo}[1][1.4em]{%
  \raisebox{-0.2\height}{\includegraphics[height=#1]{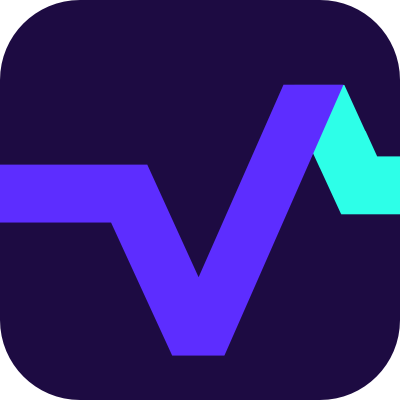}}
}

\usepackage{listings}
\usepackage{xcolor}
\definecolor{codegreen}{rgb}{0,0.6,0}
\definecolor{codegray}{rgb}{0.5,0.5,0.5}
\definecolor{codepurple}{rgb}{0.58,0,0.82}
\definecolor{backcolour}{RGB}{252, 253, 246}

\lstdefinestyle{mystyle}{
    backgroundcolor=\color{backcolour},   
    commentstyle=\color{codegreen},
    keywordstyle=\color{magenta},
    numberstyle=\tiny\color{codegray},
    stringstyle=\color{codepurple},
    basicstyle=\ttfamily\footnotesize,
    breakatwhitespace=false,         
    breaklines=true,                 
    captionpos=b,                    
    keepspaces=true,                 
    numbers=left,                    
    numbersep=5pt,                  
    showspaces=false,                
    showstringspaces=false,
    showtabs=false,                  
    tabsize=2
}

\usepackage[T1]{fontenc}

\usepackage[utf8]{inputenc}

\usepackage{microtype}

\usepackage{inconsolata}

\title{\legentlogo\hspace{2pt}UltraEval: A Lightweight Platform for \\ Flexible and Comprehensive Evaluation for LLMs}

\author{Chaoqun He$^{1}$, Renjie Luo$^{2}$\footnotemark[2], Shengding Hu$^{1}$, Yuanqian Zhao$^{3}$\footnotemark[2], Jie Zhou$^{4}$\\
\textbf{Hanghao Wu$^{4}$, Jiajie Zhang$^{5}$\footnotemark[2], Xu Han$^{1*}$,Zhiyuan Liu$^{1}$\thanks{Corresponding author:Xu Han (thu.hanxu13@gmail.com) and Zhiyuan Liu (liuzy@tsinghua.edu.cn)}}\textbf{, Maosong Sun$^{1}$}\\
\textsuperscript{1}{ Dept. of Comp. Sci. \& Tech., Institute for AI, Tsinghua University, Beijing, China}\\
\textsuperscript{2}{ Institute of Artificial Intelligence, Beihang University, China} \\
\textsuperscript{3}{ Renmin University of China} \textsuperscript{4}{ ModelBest Inc.} \textsuperscript{5}{ Northeastern University, China}\\
{\tt \{hcq21, hsd23\}@mails.tsinghua.edu.cn, renjie.luo@outlook.com}
}

\begin{document}
\maketitle

\renewcommand{\thefootnote}{\fnsymbol{footnote}}
\footnotetext[2]{Work done during internship at ModelBest Inc.}
\renewcommand{\thefootnote}{\arabic{footnote}}

\begin{abstract}

Evaluation is pivotal for refining Large Language Models (LLMs), pinpointing their capabilities, and guiding enhancements. The rapid development of LLMs calls for a lightweight and easy-to-use framework for swift evaluation deployment. 
However, considering various implementation details, developing a comprehensive evaluation platform is never easy.
Existing platforms are often complex and poorly modularized, hindering seamless incorporation into research workflows. 
This paper introduces UltraEval, a user-friendly evaluation framework characterized by its lightweight nature, comprehensiveness, modularity, and efficiency.
We identify and reimplement three core components of model evaluation (models, data, and metrics). 
The resulting composability allows for the free combination of different models, tasks, prompts, benchmarks, and metrics within a unified evaluation workflow. 
Additionally, UltraEval supports diverse models owing to a unified HTTP service and provides sufficient inference acceleration. 
UltraEval is now available for researchers publicly~\footnote{The website of UltraEval is at \url{https://github.com/OpenBMB/UltraEval} and a demo video is at \url{https://youtu.be/C0O6BVzNAS8}.}.
\end{abstract}

\section{Introduction}



LLMs have been deployed in diverse domains, such as finance\cite{zhang2023xuanyuan}, education\cite{kasneci2023chatgpt}, and law\cite{blair2023can}, demonstrating their versatility and efficacy\cite{zhao2023survey}. 
\begin{figure}[ht]
    \centering
    \includegraphics[width=0.6\linewidth]{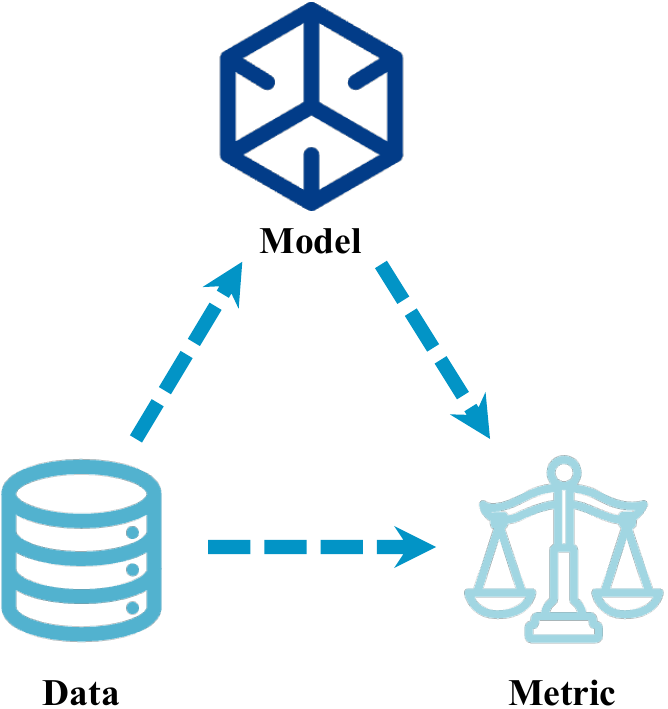}
    \caption{The three core modules of model evaluation.}
    \label{fig:fig1}
\end{figure}
This advancement is significantly bridging the gap between the realization of the current state and Artificial General Intelligence (AGI)\cite{bubeck2023sparks}. 
Nevertheless, the expansion of model parameters and training datasets engenders increasing uncertainties and emergent capabilities, posing potential risks to humanity and challenges to stable training models~\cite{chang2023survey,bommasani2021opportunities,wei2022emergent}.
Consequently, it is imperative to continuously and meticulously evaluate the evolving capabilities of LLMs throughout their development to ensure their responsible and beneficial applications.

Traditional benchmarks~\cite{zellers2019hellaswag,suzgun2022challenging,austin2021program,clark2018think} typically focus on evaluating model performance in a specific capability, making it challenging to assess the comprehensive abilities of a model. Additionally, these benchmarks generally do not include model deployment. 
Building pipelines from scratch for each combination of tasks and models for evaluation is highly inefficient and repetitive.
Therefore, an integrated evaluation framework is crucial.
\begin{figure*}[ht]
    \vspace{-1em}
    \centering
    \includegraphics[width=0.9\textwidth]{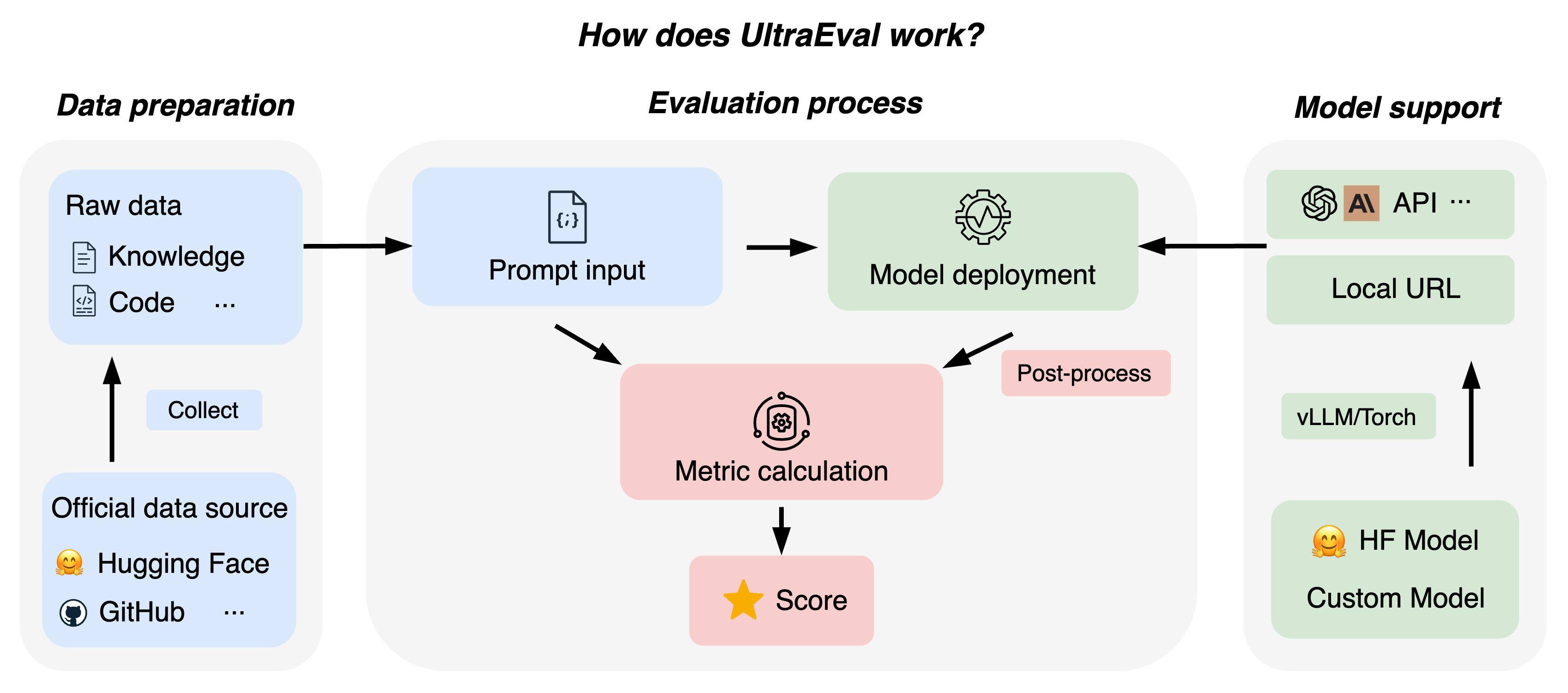}
    \caption{The overall pipeline of UltraEval, designed according to the three core modules of model evaluation.}
    \label{fig:pipeline}
\end{figure*}
Currently, some evaluation frameworks covering the entire pipeline from model deployment to model evaluation are proposed, and predominantly divided into two types: conversational websites, exemplified by platforms like Chatbot Arena~\footnote{\url{https://chat.lmsys.org/}}, and open-source evaluation tools, such as lm-evaluation-harness~\footnote{\url{https://github.com/EleutherAI/lm-evaluation-harness}}.
The former effectively assesses the conversational abilities of a model but lacks comprehensive task coverage and transparency in the evaluation process. 
The latter open-source frameworks face challenges such as incomplete task coverage, complex code structure, heavy implementations, high difficulty to use, and tightly coupled functionalities. 
These issues hinder both convenient and comprehensive assessments.

\begin{table*}[ht!]
\centering
\setlength{\tabcolsep}{6pt}
\renewcommand{\arraystretch}{1.1}
\resizebox{0.9\linewidth}{!}{
\begin{tabular}{lccccc}
\toprule
\textbf{Framework} & \textbf{Bytes} & \textbf{Datasets} & \textbf{Acceleration} & \textbf{Model Types} & \textbf{Evaluation Method}   \\ 
\midrule
Chatbot Arena      & -   & -  & -  & Chat  & Human   \\
AlpacaEval         & 3000k   & -  & -   & Chat  & GPT-4   \\
FastChat           & 950k   & 1    & -   & Chat  & GPT-4 \\
HELM               & 3200k  & 10    & -    & All   & Auto    \\
FlagEval           & 760k   & 21  & -   & All   & Auto  \\
LLM harness        & 815k  & 50+ & vLLM \& HF Accelerate & All   & Auto \\
OpenCompass        & 4000k  & 50+  & Distributed Computing & All & Auto\\
InstructEval       & 480k  & 5   & -  & Chat   & Auto  \\
OpenAI-Evals         & 780k    & -    & Concurrent API Request  & GPT  & Auto  \\
GPT-Fathom         & 445k    & 21     & Concurrent API Request   & GPT \& LLaMA  & Auto \\ 
\hline
UltraEval          & \textbf{315k}    & 50+   & vLLM \& Gunicorn   & All   & Auto \& GPT-4  \\ 
\bottomrule
\end{tabular}
}
\caption{Comparison of Evaluation Frameworks. Bytes: the total bytes of Python and Jupyter Notebook code in each framework's GitHub repository, Acceleration: tools or methods employed to expedite model inference, Model Types: Supported Models for Evaluation, GPT: GPT series models, LLaMA: LLaMA series models}
\label{tab:current_frameworks}
\end{table*}


In this paper, we identify three core components that form the evaluation process: models (or systems), task data, and metrics (i.e., evaluation methods), as illustrated in Figure~\ref{fig:fig1}. Rethinking the implementations in these three aspects would benefit the construction of a lightweight and easy-to-use evaluation framework, which covers mainstream tasks and complete evaluation pipeline, and can be easily expanded according to user customization.
To this end, we introduce UltraEval, a lightweight and user-friendly open-source framework for LLMs evaluation. It stands out for its modular and scalable design, enabling thorough assessment of model capabilities.
As illustrated in Figure~\ref{fig:pipeline}, we segment the evaluation pipeline into three main modules: Data, Model, and Metrics, each operating independently and interacting through data exchange.

Specifically, UltraEval is characterized by the following features:
\begin{enumerate}
    \item \textbf{Lightweight Usage Modes.}
    UltraEval is designed with minimal dependency requirements and features straightforward design and installation, complemented by detailed documentation. Users can initiate automated evaluations with just a few simple commands. 

    \item \textbf{Comprehensive Evaluation Tools. }
    UltraEval offers an extensive benchmarks suite, comprising over 50 commonly used benchmarks, and provides a customized prompt for each task. 
    During the evaluation process, we replicated commonly used metrics and incorporated post-processing methods for more accurate metric calculation.
    We replicate some benchmarks from the LLaMA2\cite{touvron2023llama}, achieving consistent results, which demonstrates UltraEval's reliability.

    \item \textbf{Modular Architecture and Interfaces.}
    The three main modules are independent and have clear functions, enhancing the system stability of UltraEval. Moreover, its excellent scalability allows users to flexibly customize the evaluation workflow, such as by adding new models, tasks, metrics, and more.

    \item \textbf{Efficient Inference Engines. }
    UltraEval deploys models as HTTP services, supporting the evaluation of LLMs from different sources, including the models deployed locally and the web-based API. When deployed locally, we also provide the interface to utilize vLLM~\footnote{\url{https://github.com/vllm-project/vllm}}\cite{kwon2023efficient} and Gunicorn to enable multi-GPU acceleration.
    
\end{enumerate}



Evaluation is currently in a phase of rapid and exploratory growth. 
UltraEval will be continuously updated and provide detailed tutorials to help researchers to efficiently deploy evaluation pipeline.

\section{Related Work}



The advancement of LLMs has led to the emergence of various evaluation frameworks, each with its unique features. This section will provide a detailed overview of the current state of evaluation frameworks (also see Table~\ref{tab:current_frameworks}).


\textbf{Chatbot Arena}~\cite{zheng2024judging} offers a LLM evaluation platform
where users vote on model responses, using a crowdsourced, anonymous Elo-rating system. Although innovative, its reliance on human judgment limits its suitability for fast, routine assessments. \textbf{AlpacaEval}\footnote{\url{https://github.com/tatsu-lab/alpaca_eval}}\cite{alpaca_eval} and \textbf{FastChat}\footnote{\url{https://github.com/lm-sys/FastChat}}\cite{zheng2023judging} conduct evaluation by employing GPT-4\cite{achiam2023gpt} for automated judging. 
Yet, in evaluating complex reasoning tasks, they tend to favor verbose responses and face issues with robustness. Additionally, the scope of their evaluation capabilities is limited.

\begin{table*}[ht!]
\small
\centering
\setlength{\tabcolsep}{3.8pt}
\renewcommand{\arraystretch}{1.1}
\begin{tabular}{llc}
\toprule
\textbf{First Level}       & \textbf{Second Level}           & \multicolumn{1}{c}{\textbf{Dataset List}}  \\ 
\hline
\multirow{4}{*}{Knowledge} & \multirow{3}{*}{Disciplinary knowledge}          & MMLU, CMMLU, C-Eval, AGI-Eval\\ & & JEC-QA, MEDMCQA, MEDQA-MCMLE\\& & MEDQA-USMLE, GAOKAO-Bench  \\ 
                           & World knowledge                 & NQ-open, TriviaQA, TruthfulQA  \\ \hline
Math                       & Math                            & GSM8K, MATH  \\ \hline
Code                       & Code                            & HumanEval, MBPP  \\ \hline
\multirow{5}{*}{Reason}     & Logical reasoning               & BBH  \\ 
                           & Implicative relation            & AX-B, AX-G, CB, CMNLI, OCNLI, RTE  \\ 
                           & \multirow{3}{*}{Commonsense reasoning}          & HellaSwag, OpenBookQA, ARC-c, ARC-e\\ & & CommonsenseQA, COPA, PIQA, SIQA\\  & & WinoGrande, Story Cloze, StrategyQA, TheoremQA \\ 
                           \hline
\multirow{7}{*}{Language}  & \multirow{2}{*}{Reading comprehension}           & BoolQ, C3, ChiD, DRCD, LAMBADA, MultiRC, QuAC\\ & & RACE, RECORD, SQuAD, TyDiQA, SummEdits  \\ 
                           & Translation                     & FLORES, WMT20-en-zh, WMT20-en-zh \\ 
                           & Semantic similarity             & AFQMC, BUSTM  \\ 
                           & Word sense disambiguation       & CLUEWSC, WIC, Winogender, WSC  \\ 
                           & Sentiment analysis              & EPRSTMT   \\ 
                           & News classification             & TNEWS  \\ 
\bottomrule
\end{tabular}
\caption{We compile a collection of 59 widely-used benchmarks and categorized them according to scenarios.}
\label{tab:datasets_summary}
\end{table*}

\textbf{HELM}\footnote{\url{https://github.com/stanford-crfm/helm}}\cite{liang2022holistic} streamlines language model evaluation but is constrained by its support solely for AutoModelForCausalLM\footnote{\url{https://huggingface.co/docs/transformers/main}}, excluding models without namespaces or stored locally. It lacks support for user models, demonstrates potential module coupling issues and absence of acceleration options. \textbf{FlagEval}'s\footnote{\url{https://flageval.baai.ac.cn}} ``capability-task-indicator'' framework is original but criticized for its closed-source approach and overly simplistic benchmark choices, raising data security and assessment depth concerns. Despite their innovations, both platforms fall short of the adaptability and comprehensiveness seen in more versatile frameworks like UltraEval.

\textbf{LLM harness}
\cite{eval-harness}, used by HuggingFace's Open LLM Leaderboard, and \textbf{OpenCompass}\footnote{\url{https://github.com/open-compass/opencompass}}\cite{2023opencompass} have emerged as comprehensive solutions, offering extensive dataset support and rapid updates. These feature-rich environments, however, entail a trade-off: their complexity and dependency on specific software can complicate usage and customization. This underscores the importance of detailed documentation for those looking to adapt or extend these frameworks. Similarly, \textbf{InstructEval}\footnote{\url{https://github.com/declare-lab/instruct-eval}}~\cite{chia2023instructeval}, leveraging the LLM harness infrastructure, caters specifically to models fine-tuned with instructions such as Alpaca and Flan-T5. Despite its targeted approach, InstructEval's limitations in model and task coverage hint at its niche application rather than widespread utility. The adoption of such frameworks reflects the evolving landscape of model evaluation, where finding a balance between comprehensiveness and usability poses an ongoing challenge.

\textbf{OpenAI Evals}\footnote{\url{https://github.com/openai/evals}} and \textbf{GPT-Fathom}\footnote{\url{https://github.com/GPT-Fathom/GPT-Fathom}}\cite{zheng2023gptfathom}. 
OpenAI Evals offers a straightforward, open-source framework for appraising OpenAI models, while GPT-Fathom expands upon this by analyzing the progression from GPT-3 to GPT-4 using a wider dataset array. Although it provides valuable insights into LLM development, GPT-Fathom shares OpenAI Evals' limitations in supporting a diverse range of models.


\begin{figure}[hbt!]
\vspace{-2mm}
\centering
\begin{minipage}{0.9\linewidth}
\begin{lstlisting}[language=Python]
def transform_entry(row):
    question, *choices, answer = row
    target_scores = {
        choice: int((ord(answer) - ord("A")) == idx)
        for idx, choice in enumerate(choices)
    }

    return {
        "passage": "",
        "question": question,
        "target_scores": target_scores,
        "answer": "",
    }

\end{lstlisting}
\end{minipage} 
\vspace{-0.2cm}
\caption{The data formatting template for MMLU. }
\label{fig:code_make_dataset} 
\end{figure}

\section{UltraEval}


As illustrated in Figure~\ref{fig:fig1}, evaluation is a comprehensive process that integrates models, data, and metrics. With this in mind, the design philosophy considers both the independence and interconnectivity of each module. As shown in Figure~\ref{fig:pipeline}, UltraEval encompasses the entire evaluation lifecycle\cite{chang2023survey} and segments the evaluation workflow into three main modules. In this section, we delve into the design and implementation of each component within UltraEval in detail.

\subsection{Data Preparation}
Data preparation involves transforming raw data into the final input format for the model, encompassing data preprocessing and prompt templates.

\begin{figure*}[hbt!]
\vspace{-2mm}
\centering
\begin{minipage}{0.9\linewidth}
\begin{lstlisting}[language=Python]
question = f"Question:\n{question]}\n"
instruction = f"Requirement:\nChoose and respond with the letter of the correct answer, including the parentheses.\n"
options = "Options:\n"
for idx, item in enumerate(question_options):
    options += f"({chr(65 + idx)}) {item}\n"
answer_prompt = f"Answer:\n"
final_input = question + instruction + options + answer_prompt
\end{lstlisting}
\end{minipage} 
\vspace{-0.2cm}
\caption{An example of a prompt template for an MMLU task.}
\label{fig:code_2} 
\end{figure*}

\textbf{Data Preprocessing.}
We collect commonly used benchmarks for evaluating LLMs, such as MMLU\cite{hendrycks2020measuring}, GSM8K\cite{cobbe2021training}, and Hellaswag\cite{zellers2019hellaswag}, covering multiple dimensions of capabilities. Currently, we have 59 benchmarks, listed in Table~\ref{tab:datasets_summary}, and we plan to continually expand our collection of benchmarks.

To ensure the accuracy of the data, we source it from reputable platforms like Hugging Face\footnote{\url{https://huggingface.co/datasets}} and GitHub, rather than relying on data modified by third parties. Given the varying data formats across benchmarks, we design a set of templates to standardize these diverse formats into JSON, serving as the starting point for evaluations. As shown in Figure~\ref{fig:code_make_dataset}, different data items are categorized under unified attributes.

\textbf{Prompt Templates.}
Prompts are used to guide models to generate specific outputs, and UltraEval supports prompt engineering~\cite{white2023prompt}, including few-shot and Chain of Thought (CoT)~\cite{wei2022chain}, to enhance the model's accuracy. 
The sensitivity of LLMs to prompts~\cite{zhu2023promptbench} and the variability of prompts across different tasks often make it challenging for researchers to replicate results from papers, hindering research progress. UltraEval addresses this issue by providing customized, stable prompt templates for each task to facilitate result alignment. Figure~\ref{fig:code_2} showcases an example of a prompt template for MMLU, demonstrating the rigorous process for forming the final prompt input.

\subsection{Model Deployment}
UltraEval employs a unique design approach, deploying models as HTTP online services and leveraging vLLM and gunicorn technologies to enable multi-GPU acceleration.

\textbf{Http Service.}
In traditional evaluations, model deployment is closely integrated with task assessment, requiring models to be redeployed for each new task, which can lead to unnecessary consumption of time and hardware resources.
To address this, we deploy models as HTTP services with Flask, facilitating modularization and efficient resource use. This approach has several advantages:
\begin{enumerate}
    \item \textbf{Independence.} We provide a unified interface through which models receive task data and hyperparameters, returning results upon completing inference. This setup, which allows for adjustments via hyperparameters, ensures model independence.
    \item \textbf{Comprehensiveness.} In UltraEval, we enable direct model loading from the Huggingface Transformers library. Given the independent deployment, UltraEval theoretically supports all models, including those from personal experimentation under different frameworks, greatly enhancing research and development flexibility. We utilize vLLM\footnote{\url{https://github.com/vllm-project/vllm}} as the foundational acceleration framework, granting loading priority to models it supports.
    \item \textbf{Scalability.} Thanks to its excellent scalability, users can easily extend models from language-based applications to multimodal models.
\end{enumerate}

\begin{figure*}[ht]
\vspace{-2mm}
    \centering
    \includegraphics[width=1\linewidth]{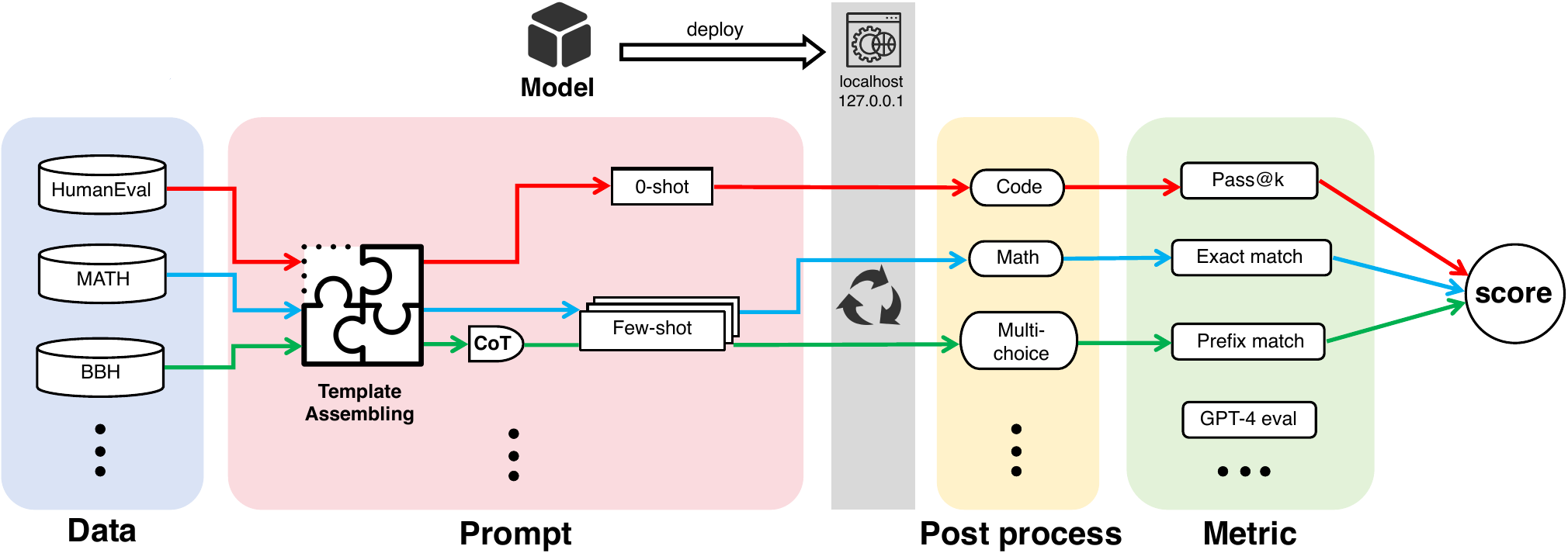}
    \caption{The combination of different modules within UltraEval.}
    \label{fig:combine}
\end{figure*}

\textbf{Multi-GPU Acceleration.} We use the Gunicorn web server with Flask to deploy models via web endpoints, achieving a flexible and decoupled architecture for model deployment and evaluation. This setup allows for dynamic GPU acceleration, where the Gunicorn server, configured with environment-specific parameters, manages multiple worker processes. Each process, handling a slice of the available GPUs, executes inference tasks in parallel, significantly improving computational efficiency and throughput. A highlight of UltraEval's performance is its ability to utilize 4 A800 GPUs to evaluate a test set of 41k data points in under 1.5 hours, showcasing remarkable efficiency\footnote{The model used in this experiment is Llama2-7B, and the benchmarks includes BBH, MMLU, C-Eval, CMMLU, HumanEval, MBPP, GSM8K and MATH. In total, they consist of 40,938 data points.}.


\subsection{Evaluation Methods}

Refined data and models are instantiated through the \textit{Task} and \textit{Model} classes, respectively, initiating the model inference process. The model performs inference based on the input data and its hyperparameters, generating prediction outcomes. Typically, 
between the model's output and the final score calculation, there are intermediate steps including post-processing and metric calculation.

\textbf{Post-Processing.}
Model outputs, influenced by task characteristics, prompt templates, and the model's performance, often contain extraneous information beyond the answers needed\cite{park2024diminished}. As shown in Figure~\ref{fig:chatgpt}, when ChatGPT responds to HumanEval questions, the response may include both code and additional textual descriptions, complicating automatic evaluation. To more accurately assess the model, it is necessary to post-process the model's outputs to extract the most crucial answers.

Post-processing is bifurcated into two dimensions: model and task. Variations in model training approaches result in different versions, such as chat and base, necessitating distinct processing methods. Additionally, certain tasks employ specific evaluation methodologies. Taking Figure~\ref{fig:chatgpt} as an example, the initial step involves extracting the code segment from ChatGPT's response. Subsequently, due to HumanEval\cite{chen2021evaluating}'s unique evaluation criteria, it is necessary to extract the function body from the code while omitting the function name, yielding a cleaner and more precise answer. UltraEval develops several post-processing methods tailored to the tasks and models currently available.

\textbf{Metric Calculation.}
Evaluation methods are categorized based on their ability to be computed automatically into automatic evaluation and human evaluation\cite{chang2023survey}. 
For automatic evaluation, we implement common metrics such as exact match for text generation tasks, F1 score for binary classification tasks, ROUGE\cite{lin-2004-rouge} for translation tasks, and pass@k\cite{chen2021evaluating} for coding tasks. Specifically, for exact match, we develop extensions like in match and prefix match to more effectively capture a wide range of scenarios.

Given that UltraEval is an automated evaluation framework, for human evaluation, we integrate GPT-4\cite{achiam2023gpt} as a discriminator to substitute for human evaluation. Moreover, all data results can be saved according to user preferences, allowing users to decide on human evaluation if desired, thus offering significant flexibility for a more objective assessment.

\begin{table}[ht!]
\centering
\resizebox{1\columnwidth}{!}{
\begin{tabular}{lcccccc}
\toprule
\multirow{2}{*}{\textbf{Benchmark}} & \multicolumn{2}{c}{\textbf{Llama2-7B}} & \multicolumn{2}{c}{\textbf{Llama2-13B}} & \multicolumn{2}{c}{\textbf{Mistral-7B}} \\ \cline{2-7} 
                           & \textbf{Official}      & \textbf{UltraEval}     & \textbf{Official}      & \textbf{UltraEval}      & \textbf{Official}      & \textbf{UltraEval}      \\ \hline
ARC-C                      & 45.9          & 43.2         & 45.9          & 47.4          & 55.5          & 50.8          \\
HellaSwag                  & 77.2          & 75.6         & 80.7          & 79.1          & 81.3          & 80.4          \\
BBH                        & 32.6          & 32.8          & 39.4          & 39.2          & $38.0^*$            & 40.4          \\
MATH                       & 2.5           & 2.8          & 3.9           & 4.8           & 13.1          & 10.2          \\
GSM8K                      & 14.6          & 14.8         & 28.7          & 22.6          & $52.1^*$          & 31.9          \\
HumanEval                  & 12.8          & 12.8          & 18.3          & 17.1          & 30.5          & 26.8          \\
MBPP                       & 20.8          & 20.8         & 30.6          & 29.0          & 47.5          & 47.3          \\
MMLU                       & 45.3          & 45.1         & 54.8          & 55.2          & 60.1          & 63.1          \\ 
\bottomrule
\end{tabular}
}
\caption{Evaluation results on mainstream benchmarks (\%). $^*$ The BBH score is not explicitly stated in the paper\cite{jiang2023mistral}, however, it is inferred to be 38.0 from the figures in the paper. The replicated result for GSM8K is 35.4 in Gemma paper\cite{team2024gemma}, which is close to our result.}
\label{tab:eval result}
\end{table}

\section{Evaluation}
UltraEval aims to provide a lightweight, comprehensive, and user-friendly evaluation framework to support research. As illustrated in Figure~\ref{fig:combine}, UltraEval's modular design effectively combines various models, tasks and metrics for evaluation. Using UltraEval, we evaluate models from the LLaMA2 series\cite{touvron2023llama} and Mistral\cite{jiang2023mistral} on these widely-used benchmarks. As indicated in the Table~\ref{tab:eval result}, 
some reproduced results are higher, while others are lower, but within a certain margin of error, our reproduced results are consistent with the results reported in the papers, underscoring our framework's reliability. The sources of error include hyperparameters (e.g., temperature, top-p) and hardware configurations. Since the evaluation details are not provided in the papers, verification is not possible. This highlights the importance of having an open and reproducible evaluation framework.

Furthermore, UltraEval supports innovative research efforts, research on predictable scaling\cite{hu2023unlock}, OlympiadBench\cite{he2024olympiadbench} and model training, such as with MiniCPM\cite{hu2024minicpm}.

\section{Discussion and Future Work}
In this section, we discuss future work following this study. Specifically, we focus on addressing data contamination and supporting a broader range of evaluation scenarios.

\textbf{Data contamination} refers to the phenomenon that examples from the evaluation set are also found in the training data~\cite{li2024latesteval}, causing inaccuracies in model evaluation. Common methods for contamination detection include n-gram overlap and embedding similarity search. However, none of these methods are perfect, making research on contamination detection still crucial. We will integrate these methods.

\textbf{Supporting more evaluation scenarios}, such as multimodal, long-text, and Retrieval-Augmented Generation (RAG), is crucial for meeting a broader range of evaluation needs. This will be an important direction for our future work.

\section{Conclusion}
We introduce UltraEval, a lightweight, user-friendly, and comprehensive framework for model evaluation. UltraEval establishes a unified structure with well-defined modules and flexible interactions, aiding researchers and developers in efficiently deploying evaluation workflows. Moving forward, we plan to continuously integrate new technologies and features into UltraEval, extending beyond large language models to support the evaluation of multimodal models, Retrieval-Augmented Generation (RAG), agents, and more, to advance the research on AGI.
Additionally, we aim to expand our collection of representative benchmarks and also develop our own, exploring the capabilities and limits of large models.


\section*{Acknowledgements}
This work is supported by the National Key R\&D Program of China (No. 2022ZD0116312), Quan Cheng Laboratory (Grant No. QCLZD202301), the Postdoctoral Fellowship Program of CPSE (Grant No. GZB20230343), China Postdoctoral Science Foundation (Grant No. 2023M741945), and Institute Guo Qiang at Tsinghua University.

\section*{Limitations}
Currently, our approach primarily utilizes text domain evaluation. However, we are looking to expand the scope of UltraEval by integrating multi-modal and long-context evaluation datasets. This enhancement aims to facilitate more thorough and diverse assessments. Additionally, there is room for improvement in the visualization of our results. Future improvement will focus on enabling multi-dimensional visualization, thereby enriching the interpretability and depth of our evaluation results.

\section*{Ethical Considerations}
In this paper, we present UltraEval, a lightweight, user-friendly, flexible, and comprehensive framework for model evaluation. Adhering to the principles of modularity, UltraEval segments the evaluation process into three distinct modules: Data, Model, and Metrics. This approach enhances the framework's extensibility and flexibility, allowing for the easy integration of new models and tasks. We offer an extensive benchmark suite and replicate commonly used models and benchmarks. Our results align with those reported in the corresponding papers, underscoring the stability and reliability of our framework. Committed to sustainable development, we publicly release all our code to minimize unnecessary carbon footprint. Throughout our experiments, we adhere to all licenses related to models and data.

\bibliography{custom}

\appendix

\section{More details}
\label{sec:appendix}

\subsection{Post-process}

We present the post-processing code in Figure~\ref{fig:code_process} and explain the reasons necessitating post-processing in Figure~\ref{fig:chatgpt}.

\subsection{UltraEval Usage}
In the tutorial~\footnote{\url{https://github.com/OpenBMB/UltraEval/blob/main/docs/tutorials/en/ultraeval.md}}, we provide a detailed guide on UltraEval, including an introduction to its modules and instructions on how users can customize evaluations, such as adding their own new tasks and new models. This ensures that diverse evaluation needs are met.

\subsection{Model as Judge}
Currently, UltraEval supports ChatGPT as a substitute for human evaluation. However, since ChatGPT is a commercial model, users need to provide the relevant API key. Due to the modularity and flexibility of our framework, users can also use other closed-source models or highly performant open-source models as evaluators.

\subsection{Multi-GPU Acceleration}
Multi-GPU acceleration requires sufficient GPU resources. For larger models, such as Llama2-70B, deploying a single instance requires two A100 GPUs. Therefore, when using Multi-GPU acceleration, it is essential to consider both the model and the available GPU resources to ensure optimal configuration.


\begin{figure*}[hbt!]
\vspace{2mm}
\centering
\begin{minipage}{0.9\linewidth}
\begin{lstlisting}[language=Python]

def process_text(text):
    triple_quotes_indices = [i for i, _ in enumerate(text) if text.startswith('\"\"\"', i)]

    if len(triple_quotes_indices) % 2 == 0 and len(triple_quotes_indices) > 0:
        for i in range(0, len(triple_quotes_indices), 2):
            start_index = triple_quotes_indices[i]
            end_index = triple_quotes_indices[i + 1]
            if 'def' in text[end_index:]:
                return text[:start_index].strip()
        return text.strip()
    elif len(triple_quotes_indices) > 0:
        return text[:triple_quotes_indices[0]].strip()
    else:
        return text.strip()

\end{lstlisting}
\end{minipage} 
\vspace{-0.2cm}
\caption{An example of a post-processing method for the MBPP task.}
\vspace{-0.5cm}
\label{fig:code_process} 
\end{figure*}



\begin{figure*}[ht]
    \vspace{-1em}
    \centering
    \includegraphics[width=1\textwidth]{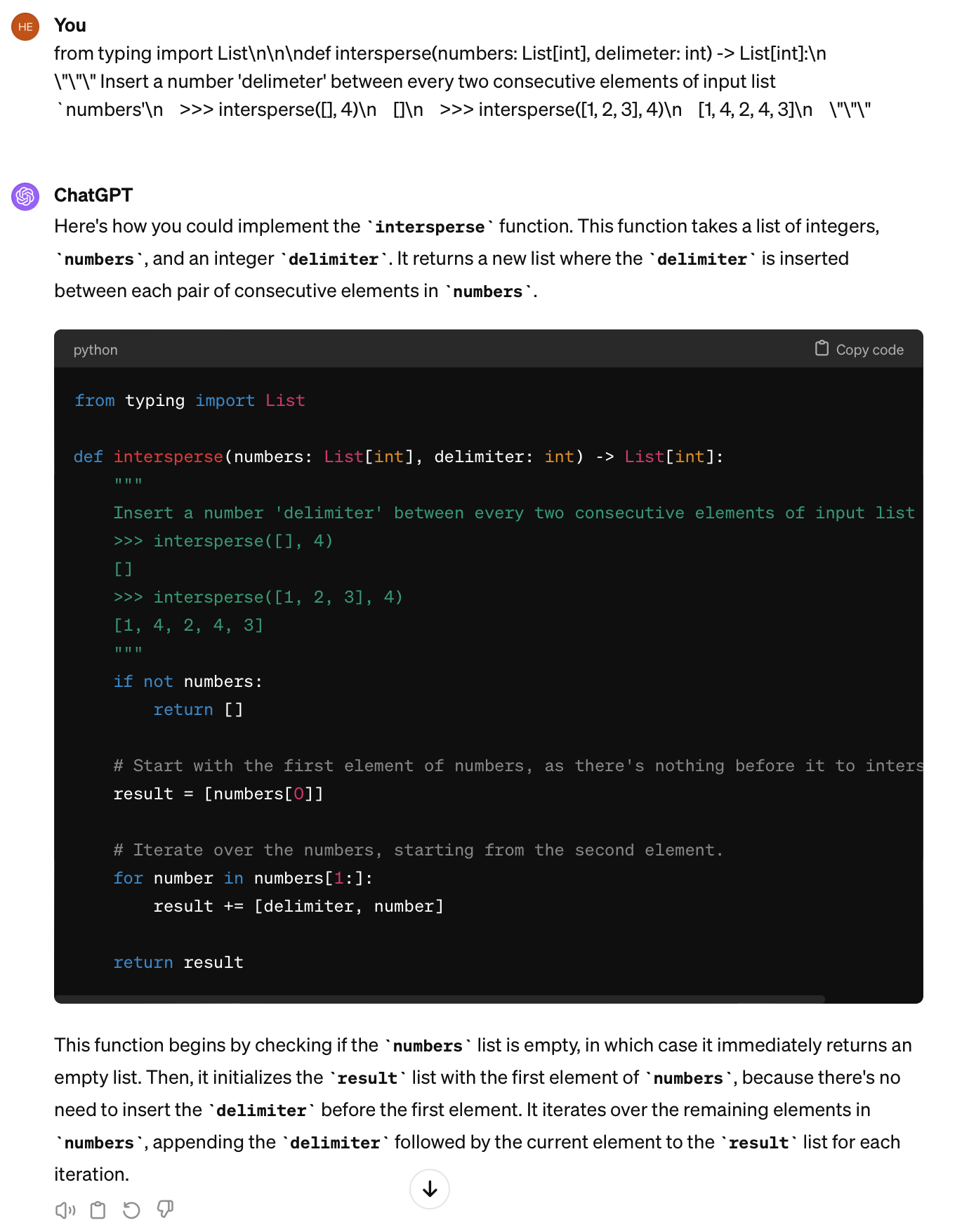}
    \caption{An example of ChatGPT to accomplish HumanEval. The figure illustrates that the responses from ChatGPT cannot be directly used for computation and require post-processing to extract the substantive content.}
    \label{fig:chatgpt}
\end{figure*}

\end{document}